\documentclass[11pt,a4paper]{article}
\usepackage[hyperref]{acl2020}
\usepackage{times}
\usepackage{latexsym}

\usepackage{microtype}

\usepackage{booktabs}
\usepackage{amsbsy,amssymb}
\newcommand*\samethanks[1][\value{footnote}]{\footnotemark[#1]}
\usepackage{multirow}
\usepackage{multicol}
\usepackage{graphicx}
\usepackage{adjustbox}
\usepackage{CJKutf8}
\usepackage{bm}

\aclfinalcopy 

\title{
Keyframe Segmentation and Positional Encoding \\
for Video-guided Machine Translation Challenge 2020
}

\author{
        Tosho Hirasawa \thanks{ \ \ Equal contribution}\\
         Tokyo Metropolitan \\  University \\
        \texttt{hirasawa-tosho} \\ \texttt{@ed.tmu.ac.jp}
        \And 
        Zhishen Yang \samethanks \\
        Tokyo Institute \\ of Technology \\
        \texttt{zhishen.yang} \\ \texttt{@nlp.c.titech.ac.jp}
        \And 
        Mamoru Komachi \\
        Tokyo Metropolitan \\ University \\
        \texttt{komachi} \\ \texttt{@tmu.ac.jp}\\
        \And Naoaki Okazaki \\
        Tokyo Institute  \\of Technology\\
        \texttt{okazaki} \\ \texttt{@c.titech.ac.jp}
}

\date{}

\begin{document}
\maketitle
\begin{abstract}

Video-guided machine translation as one of multimodal neural machine translation tasks targeting on generating high-quality text translation by tangibly engaging both video and text. In this work, we presented our video-guided machine translation system in approaching the Video-guided Machine Translation Challenge 2020. This system employs keyframe-based video feature extractions along with the video feature positional encoding. In the evaluation phase, our system scored 36.60 corpus-level BLEU-4 and achieved the 1st place on the Video-guided Machine Translation Challenge 2020.

\end{abstract}

\section{Introduction}

In multimodal machine translation (MMT), a target sentence is translated from a source sentence together with related nonlinguistic information such as images \cite{specia2016shared} and videos \cite{wang2019vatex}. The goal of Video-guided Machine Translation (VMT) Challenge 2020 is to generate target-language video descriptions given both the videos and its description in the source languages.

Videos preserve rich visual information that guides textual translation. Since video descriptions illustrate visual objects, scenes and actions in the videos, we hypothesize that obtaining appearance features (objects and scenes) and action features from videos will contribute to the quality of translation. Additionally, keyframes store entire images in the video, helping us extract high-quality video features.

A video consists of an ordered sequence of frames, while features extracted from them often do not preserve such order information. We hypothesize that incorporating such order information with visual feature facilities our model in improving translation quality.



Based on the above hypotheses, in the proposed video-guided machine translation system, we introduce a keyframe-based approach for video feature extraction, along with positional encoding to inject order information into video features. The core model in our system is a modified hierarchical attention model \cite{libovicky2017attention} with encoder and decoder architecture.

\section{Hierarchical Attention with Positional Encoding}

Our video-guided machine translation system is an extension of the hierarchical attention model \cite{libovicky2017attention}. The underlying model has a simple encoder and a modified decoder from \newcite{bahdanau2015neural} that uses two individual attention mechanisms to compute the textual context vector and the auxiliary context vector (in our case, the context vector over sequential video representations). However, the model is assumed to incorporate with spatial image features (e.g., region of Interest feature from Faster-RCNN models) and cannot leverage order information, which is a distinguishing property of video features (e.g. I3D).

To address this problem, we add positional encodings \cite{vaswani2017attention} to the video representations at the beginning of the attention to make the model use the order of the representations.


\paragraph{Encoder} 
We first encode the $N$-tokens input sentence $\textbf{x}=(x_1,\cdots,x_N)$ into encoder states $\textbf{h}=(h_1,\cdots,h_N)$ by a bidirectional GRU, where each $h_*$ is a vector with $d$ dimension.
The $T$-elements video representations $\textbf{z}=(z_1,\cdots,z_T)$ is extracted from a video $\bm{v}$ using either video or imagery encoder described in Section \ref{vis_enc_sec}. 

Additionally, we add positional encoding to video representations $\textbf{z}$ to obtain position-aware video representations $\hat{\textbf{z}} = (\hat{z}_1, \cdots, \hat{z}_T)$ at each timestep $pos\in(1, \cdots, T)$:
\begin{eqnarray}
    &\hat{z}_{pos}   &= z_{pos} + PE_{pos} \\
    &PE_{(pos,2i)}   &= sin(pos/10000^{2i/d}) \\
    &PE_{(pos,2i+1)} &= cos(pos/10000^{2i/d})
\end{eqnarray}
where $i$ is the dimension.

\paragraph{Decoder}
In each position $j$ while decoding, we first compute the decoder state proposal $\bm{s}_j$ from previous word embedding $w_{j-1}$ and previous decoder state $\hat{\bm{s}}_{j-1}$,
\begin{equation}
    \bm{s}_j = \mathrm{GRU}(w_{j-1}, \hat{\bm{s}}_{j-1})
\end{equation}

Afterward, the textual context vector $\bm{c}_j^{(t)}$ and the video context vector $\bm{c}_j^{(\bm{z})}$  are computed using two separate attention mechanisms $\mathrm{att}_{\bm{t}}$ and $\mathrm{att}_{\bm{z}}$.
\begin{eqnarray}
  \bm{c}_j^{(\bm{t})} = \mathrm{att}_{\bm{t}}(\bm{s}_j, \textbf{h}) \\
  \bm{c}_j^{(\bm{z})} = \mathrm{att}_{\bm{z}}(\bm{s}_j, \hat{\textbf{z}})
\end{eqnarray}

The final context vector $\bm{c}_j$ is computed using another attention over modalities $m\in\{\bm{t},\bm{z}\}$.
\begin{eqnarray}
    & \bm{e}_{j}^{(m)} &= o^{T} \mathrm{tanh}(\bm{W}_1 s_j + \bm{U}^{(m)} \bm{c}_j^{(m)}) \\
    & \alpha_j^{(m)} &= \frac{
        \mathrm{exp}(\bm{e}_{j}^{(m)})
    }{
        \sum\limits_{m'\in \{t, z\}} \bm{e}_{j}^{(m')}
    } \\
    & \bm{c}_j &= \sum\limits_{m\in \{t, z\}} \alpha_j^{(m)} \bm{Q}^{(m)} \bm{c}_j^{(m)}
\end{eqnarray}
where $o^{T}$ and $\bm{W}_1$ are model parameters and shared among all modalities, while $\bm{U}^{(m)}$ and $\bm{Q}^{(m)}$ are dedicated model parameters for each modality. 
$\bm{U}^{(m)}$ and $\bm{Q}^{(m)}$ are a projection matrices that map each single-modality context vector into a common space. 
$o$ is the weight vector with the same dimensions of the common space.

The final context vector $\bm{c}_j$ is fed into the second GRU along with the decoder state proposal $\bm{s}_j$ to generate the final decoder state $\hat{\bm{s}}_j$ and output distribution $p(y_j|y_{<j})$
\begin{eqnarray}
    &\hat{\bm{s}}_j &= \mathrm{GRU}(\bm{c}_j, \bm{s}_j) \\
    &p(y_j|y_{<j})  &= \mathrm{softmax}(\bm{W}_2 \hat{\bm{s}}_j + b)
\end{eqnarray}
where $\bm{W}_2$ and $b$ are model parameters.

\paragraph{Multiple Video Feature Integration}

Integrating various types of video features into a video-guided machine translation system represents a potential way of improving translation quality \cite{wang2018watch}. In our system, we decided to ensemble models trained on different types of video features. In section~\ref{experimental_result}, we detail choices on ensemble models.


\section{Video and Imagery Encoders}
\label{vis_enc_sec}

Videos, as another input in our system, often possess visual clues that guide the translation, such as actions, objects and scenes. Encoding video to acquire information-rich video features acts as visual-guidance to text translation.

We classified two types of video features: action features derived from actions, and appearance features from visual objects and scenes. Keyframes in videos store whole images, which often provide good visual representations of objects and scenes. We used keyframes for appearance feature extraction and as a basis to segment videos for obtaining motion features.

Our video-guided machine translation system consists of a video encoder that outputs action features, and an imagery encoder generates appearance features.

\begin{table*}[th]

  \begin{center}
  \begin{adjustbox}{width=0.9\linewidth}
    \begin{tabular}{l|c|c}
      \toprule
        Model                                   &  Validation Set & Public Test Set \\
      \midrule[0.08em]
        \newcite{wang2019vatex}                 & -     & 29.12 \\
      \midrule[0.08em]
        (1) Text-only                            & 35.10 & -  \\
        \midrule[0.08em]
        Official I3D features  ~\footnote{\url{https://eric-xw.github.io/vatex-website/download.html}} & & \\
        \hspace{2em} (2) with positional encoding   & 35.28 & 35.26 \\
        \hspace{2em} (3) without positional encoding    & 35.02 & -     \\
        \midrule[0.08em]
        Keyframe-based video feature extraction                     &       &       \\
        \hspace{2em} (3) Action features                          & \textbf{35.42} & \textbf{35.35} \\
        \hspace{2em} (4) Object features (Res-Net 152, ImageNet)              & 35.29 & -     \\
        \hspace{2em} (5) Scene features (Res-Net 50, Place365)                & 35.14 & -     \\

        \midrule[0.08em]
       Ensemble Model                              &       &       \\
          3 Action features (3)                                 & 36.20 & -     \\
          3 Object features (4) + 3 Scene features (5)               & 36.38 & -     \\
          3 Action features (3) + 3 Object features (4) + 3 Scene features (5)       & \textbf{36.48} & \textbf{36.60} \\
      \bottomrule
    \end{tabular}{}
 \end{adjustbox} 
  \end{center}

  \caption{
    Corpus-level BLEU on validation and public\_test sets.
  }
  \label{main_result}
\end{table*}

\paragraph{Video Encoder} 

We first segmented a video based on keyframes \footnote{https://github.com/dmlc/decord} to build a segment list, and each segment contains a keyframe and 31 consecutive frames after it. 
We then feed the video segment list to obtain the action feature matrix from a non-local neural network~\cite{wang2018non} with Res-Net 101~\cite{he2016deep} as the backbone pre-trained on ImageNet and fine-tuned on Kinetics400 dataset~\footnote{https://gluon-cv.mxnet.io/model\_zoo/action\_recognition.html\#id113}. 
Each feature vector in the matrix is in chronological order of appearance of its video segment by the time. 
The action feature matrix $\textbf{M} \in \mathbf{R}^{T \times d}$ for a video $\bm{v}$ is:
\begin{eqnarray}
    &\textbf{M}&= \mathrm{Video \ Encoder}(S) \\
    &S &= \mathrm{Segmentation}(\bm{v})
\end{eqnarray}
where $S$ is the list of $T$ keyframe video segments with chronological order. 

\paragraph{Imagery Encoder} 
Keyframes in the video store complete imagery information that suites our need to extracting high-quality appearance features. Frames in the video often involve visual objects and visual scenes. 
Therefore we obtained these two types of appearance features from keyframes. 
We employed a object-recognition system that is a Res-Net 152 model pre-trained with ImageNet  and a scene-recognition system that is a Res-Net 50 model pre-trained with Place365~\cite{zhou2017places}~\footnote{\url{http://places2.csail.mit.edu/}}.
The input to our imagery encoder is the keyframes from the video.

\section{Experiment setup}

\paragraph{Model}

The encoders of our model have one layer with 512 hidden dimensions, and therefore the bidirectional GRU has a dimension of 1024. The decoder state has a dimension of 512. The input word embedding size and output vector size are 1024.

During training, we used Adam optimizer with a learning rate of 0.001, clipping gradient norm to 1.0, the dropout rate of 0.5, batch size of 512, and early stopping patience of 10. The loss function was cross entropy. In the evaluation phase, we performed a beam search with a size of 5.

\paragraph{Preprocess}

We preprocessed both English and Chinese sentences in the same manner as in the starter code \footnote{https://github.com/eric-xw/Video-guided-Machine-Translation}, where English sentences are lower-cased, and Chinese sentences are split into sequences of characters.

The vocabulary of either English or Chinese contains tokens that occur at least five times in the training set, giving 7,947 types for English and 2,655 types for Chinese.

\section{Experimental Result}
\label{experimental_result}
The official evaluation metrics for VMT challenge 2020 is BLEU~\cite{papineni2002bleu}. Table \ref{main_result} shows the corpus-level BLEU-4 scores of each model on the validation set and public test set.

\begin{figure*}[th]
    \centering
    \includegraphics[width=\linewidth]{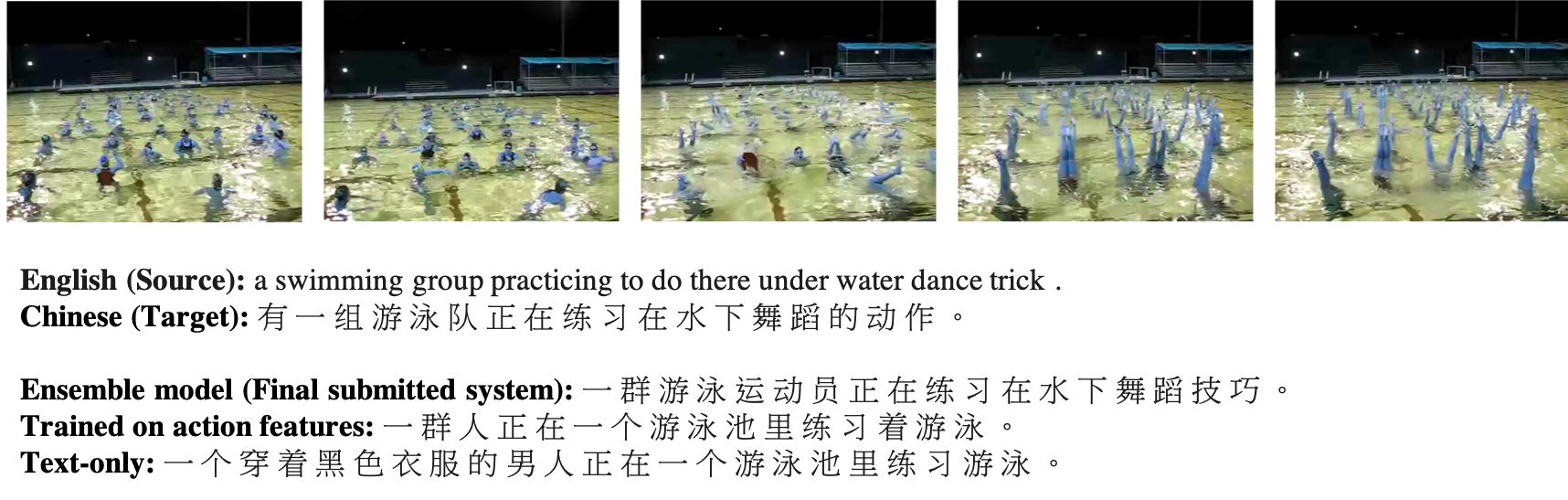}
    \caption{Example translations from the validation set generated from three of our system variants. Only our system with ensemble model (final submission) could correctly translate the group people as \begin{CJK}{UTF8}{gbsn}游泳运动员\end{CJK} (swimming athletes), and their actions as \begin{CJK}{UTF8}{gbsn}练习水下舞蹈技巧\end{CJK} (practicing underwater dance tricks).}
    \label{example}
\end{figure*}

Our model using textual feature achieved a score of 35.10 on validation set, which makes it serve as a baseline (text-only baseline) for the following experiments with inputs of both textual and video features. Engaging official video features, we observed slight performance deterioration (-0.08 BLEU) without positional encoding, while a score improvement (+0.18 BLEU) with it. Based on above results, we decided to keep positional encoding in our system.

We extracted two types of video features based on keyframes: action features and appearance features, in which appearance features consist of two types of features: object features, and scene features. To evaluate how each type of video feature helps in improving the translation quality, we trained our system using each type of video feature with textual features.

On evaluation set, training our system using object features ((4) in the table~\ref{main_result}) and textual features improves 0.19 BLEU scores over text-only baseline, and 0.01 BLEU compared to system trained on official I3D features ((2) in table~\ref{main_result}). Training on scene features ((5) in table ~\ref{main_result}) and textual features give the system a slight 0.04 BLEU score improvements over text-only baseline. Compared to the appearance features, our system trained on action features and textual features achieved the highest BLEU scores, particularly, 0.32 BLEU over text-only baseline and 0.14 BLEU over our system trained on official I3D features. Based on above results, all types of video features improve our system performance compared to using textual feature only.

Compared to our system trained on single types of video feature and textual feature, our system ensembling models trained on different types of video features and textual features give another raise in the BLEU score. On evaluation set, compared to best preform single video feature model (3 in table ~\ref{main_result}), ensemble three models of (3) improves 0.78 BLEU score, while ensemble 3 models of (4) and 3 models of (5) get 0.96 BLEU score boost.  An ensemble of three different models (3), (4), and (5) achieves the best BLEU score on validation set, this ensemble model is also our final submission, which obtains 36.60 BLEU score in the public test set an ranks the first place. Figure~\ref{fig:my_label} shows example translations generated from three of our system variants.

\section{Conclusion}

In the Video-guided Machine Translation Challenge 2020, we revealed that keyframe-based video feature extraction and positional encoding jointly enhance the translation quality by showing a substantial improvement from the text-only baseline.

We also demonstrated that the ensemble of multiple models trained on different types of video features brought further performance improvements. In the future, we will explore the best integration of different features to improve translation quality under the video-guidance.

\footnotesize
\bibliography{anthology,acl2020}
\bibliographystyle{acl_natbib}

\end{document}